\documentclass[10pt,twocolumn,letterpaper]{article}

\usepackage{iccv}
\usepackage{times}
\usepackage{epsfig}
\usepackage{graphicx}
\usepackage{amsmath}
\usepackage{amssymb}

\usepackage{multirow}
\usepackage{array}
\usepackage{booktabs}
\usepackage{subcaption}
\usepackage[table,xcdraw]{xcolor}
\usepackage{macro}
\usepackage{float}
\newcommand{\nrsfm}{NRS\emph{f}M\xspace}
\newcommand{\sfc}{S\emph{f}C\xspace}
 \renewcommand{\subsubsection}[1]{\vspace{5pt} \noindent \textbf{#1:}}

\usepackage[pagebackref=true,breaklinks=true,letterpaper=true,colorlinks,bookmarks=false]{hyperref}

\iccvfinalcopy 

\def\iccvPaperID{3127} 

\ificcvfinal\fi
\begin{document}

\title{Deep Non-Rigid Structure from Motion}

\author{Chen Kong\\
Carnegie Mellon University\\
{\tt\small chenk@cs.cmu.edu}
\and
Simon Lucey\\
Carnegie Mellon University\\
{\tt\small slucey@cs.cmu.edu}
}

\maketitle

\begin{abstract}
Current non-rigid structure from motion (NRSfM) algorithms are mainly limited with
 respect to: (i) the number of images, and (ii) the type of shape variability
 they can handle. This has hampered the practical utility of NRSfM for many
 applications within vision. In this paper we propose a novel deep neural
 network to recover camera poses and 3D points solely from an ensemble of 2D
 image coordinates. The proposed neural network is mathematically interpretable
 as a multi-layer block sparse dictionary learning problem, and can handle
 problems of unprecedented scale and shape complexity. Extensive experiments
 demonstrate the impressive performance of our approach where we exhibit
 superior precision and robustness against all available state-of-the-art works
 by an order of magnitude.  We further propose a quality measure (based on 
 the network weights) which circumvents the need for 3D ground-truth to 
 ascertain the confidence we have in the reconstruction. 
\end{abstract}

\section{Introduction}
\vspace{-0.25cm}
Building an AI capable of inferring the 3D structure and pose of an object from
a single image is a problem of immense importance. Training such a system using
supervised learning requires a large number of labeled images -- how to obtain
these labels is currently an open problem for the vision community.
Rendering~\cite{su2015render} is problematic as the synthetic images seldom
match the appearance and geometry of the objects we encounter in the
real-world. Hand annotation is preferable, but current strategies rely on
associating the natural images with an external 3D dataset (\eg
ShapeNet~\cite{DBLP:journals/corr/ChangFGHHLSSSSX15}, ModelNet~\cite{wu20153d}),
which we refer to as \emph{3D supervision}. If the 3D shape dataset does not
capture the variation we see in the imagery, it is inherently
ill-posed.

\begin{figure}[t]
 \centering
 \includegraphics[width=\linewidth]{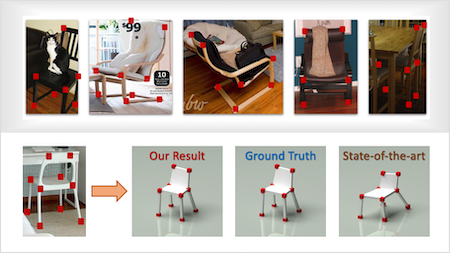}
 \caption{In this paper, we want to reconstruct 3D shapes solely from a sequence of
 annotated images---shown on top---with no need of 3D ground truth. Our proposed
 hierarchical sparse coding model and corresponding deep solution outperforms state-of-the-art by an order of magnitude. }
 \label{fig:teaser}
\end{figure}

Non-Rigid Structure from Motion (\nrsfm) offers computer vision a way out of
this quandary -- by recovering the pose and 3D structure of an object category
\emph{solely} from hand annotated 2D landmarks with no need for 3D supervision.
Classically~\cite{bregler2000recovering}, the problem of \nrsfm has been applied
to objects that move non-rigidly over time such as the human body and face. But
\nrsfm is not restricted to non-rigid objects; it can equally be applied to
rigid objects whose object categories deform non-rigidly~\cite{kong2016sfc, agudo2018image, vicente2014reconstructing}.
Consider, for example, the five objects in Figure~\ref{fig:teaser}~(top), instances from
the visual object category ``chair''. Each object in
isolation represents a rigid chair, but the set of all 3D shapes describing
``chair'' is non-rigid. In other words, each object instance can be modeled as
a deformation from its category's general shape.

\nrsfm is well-noted in literature as an ill-posed problem due to the non-rigidity. 
This has been mainly addressed by imposing additional shape priors, \eg 
low rank~\cite{bregler2000recovering, dai2014simple}, 
union-of-subspaces~\cite{zhu2014complex, agudo2018image}, 
and block-sparsity~\cite{kong2016prior, kong2016sfc}. However, low rank is only applicable to
simple non-rigid objects with limited deformations and union-of-subspaces relies heavily on frame
clustering which is still an open problem. Block-sparsity, where each shape can be represented by at 
most $K$ bases out of $L$, is considered as one of the most promising assumptions in terms of modeling
broad shape variations. This is because sparsity can be thought of as a union of 
$\binom{L}{K}$ subspaces constraint and an over-complete dictionary could be utilized. 
However, it was noted by Kong~\etal~\cite{kong2016prior}, that searching for the best subspace out of 
$\binom{L}{K}$ is extremely costly and not robust. Based on this observation we propose a novel shape prior that employs hierarchical sparse coding. The introduction of additional layers, in comparison to classical single-layer sparse coding, provides a mechanism for controlling the number of active subspaces. The ability of hierarchical sparse coding to provide a model that is both expressible and robust for general \nrsfm forms the central thesis of our paper.

\subsubsection{Contributions}
\begin{packed_item}
    \item We propose a novel shape prior based on hierarchical sparse coding and 
    demonstrate that the 2D projections under orthogonal cameras can be 
    represented by the hierarchical dictionaries in a block sparse way.
    \item We designed a deep neural network to approximately solve the proposed
    hierarchical block sparse model and show how the network architecture is derived 
    from a classical sparse coding algorithm.
    \item Finally, extensive experiments are conducted using various datasets.
    Both quantitative and qualitative results demonstrate our superior performance,
    outperforming all state-of-the-arts in the order of magnitude.
\end{packed_item}

\section{Related Work}
\vspace{-0.25cm}
\subsubsection{Low-rank \nrsfm}
In rigid structure from motion, the rank of 3D structure is fixed as three~\cite{tomasi1992shape}
since 3D shapes remain the same between frames. Based on this insight, Bregler~\etal~\cite{bregler2000recovering} advocated that non-rigid 3D structure could be represented by a linear subspace of low rank. Dai~\etal~\cite{dai2014simple} developed this prior by proving that low-rank assumption itself is sufficient to address the ill-posedness of
\nrsfm with no need of additional priors. The low-rank assumption has also been applied temporally~\cite{akhter2009defense, fragkiadaki2014grouping} -- 3D point trajectories can be represented by pre-defined (\eg~DCT) or learned bases. Although exhibiting impressive performance, the low-rank assumption has a major drawback. The rank is strictly 
limited by the number of points and frames (whichever is smaller~\cite{dai2014simple}). This makes low-rank \nrsfm~infeasible if we want to solve large-scale problems with complex shape variations when the number of points is substantially smaller than the number of frames. 

\subsubsection{Union-of-subspaces \nrsfm}
Inspired by an intuition that complex non-rigid deformations could be clustered into
a sequence of simple motions, Zhu~\etal~\cite{zhu2014complex} proposed to model non-rigid
3D structure by a union of local subspaces. This was later extended to spatial-temporal domain~\cite{agudo2017dust} and applied to rigid object category reconstruction~\cite{agudo2018image}.
The major difficulty of union-of-subspaces is how to effectively cluster shape
deformations purely from 2D observations and how to estimate affinity matrix when the number
of frame is large \eg more than tens of thousand frames. 

\subsubsection{Sparse \nrsfm}
Sparse prior~\cite{kong2016prior, zhou2015sparseness, kong2016sfc} is more generic than
union-of-subspaces since it is equivalent to the union of all possible local subspaces. One
obvious advantages of this is the large number of subspaces enables the effective modeling of a broader set of 3D
structures. However, the sheer number of subspaces that can be entertained by the sparsity prior is its fundamental drawback. Since there are so many possible subspaces to choose from, the approach is sensitive to noise, dramatically limiting its applicability to ``real-world'' \nrsfm problems. In this paper we want to leverage the elegance and expressibility of the sparsity prior without suffering from its inherent sensitivity to noise.

\section{Background}
\vspace{-0.25cm}
\label{sec: scdnn}
Sparse dictionary learning can be considered as an unsupervised learning task
and divided into two sub-problems: (i) dictionary learning, and (ii)
sparse code recovery. Let us consider sparse code recovery problem, where we
estimate a sparse representation $\zv$ for a measurement vector $\xv$ given the dictionary~$\Dv$
\ie
\begin{equation}
 \min_\zv \Vert \xv - \Dv\zv\Vert_2^2 \quad \st \Vert \zv \Vert_0 < \lambda,
 \label{eq:sparse_coding}
\end{equation}
where $\lambda$ related to the trust region controls the sparsity of recovered
code. One classical algorithm to recover the sparse representation is
Iterative Shrinkage and Thresholding Algorithm
(ISTA)~\cite{daubechies2004iterative, rozell2008sparse,beck2009fast}. ISTA
iteratively executes the following two steps with $\zv^{[0]} = \zero$:
\begin{gather}
 \vv = \zv^{[i]} - \alpha\Dv^T(\Dv\zv^{[i]} - \xv), \\
 \zv^{[i+1]} = \argmin_{\uv} \frac{1}{2}\Vert \uv - \vv \Vert^2_2 + \tau\Vert\uv\Vert_1,
\end{gather}
which first uses the gradient of~$\Vert \xv - \Dv\zv\Vert_2^2$ to
update~$\zv^{[i]}$ in step size $\alpha$ and then finds the closest sparse
solution using an $\ell_1$ convex relaxation. It is well known in literature
that the second step has a closed-form solution using the soft thresholding operator.
Therefore, ISTA can be summarized as the following recursive equation:
\begin{equation}
 \zv^{[i+1]} = h_\tau\big(\zv^{[i]} - \alpha\Dv^T(\Dv\zv^{[i]} - \xv)\big),
 \label{eq:ista}
\end{equation}
where $h_\tau$ is a soft thresholding operator and $\tau$ is related to
$\lambda$ for controlling sparsity.

Recently, Papyan~\cite{papyan2017convolutional} proposed to use ISTA and sparse
coding to reinterpret feed-forward neural networks. They argue that feed-forward
passing a single-layer neural network~$\zv = \relu(\Dv^T\xv - b)$ can be
considered as one iteration of ISTA when~$\zv~\ge~0, \alpha=1$ and~$\tau = b$.
Based on this insight, the authors extend this interpretation to feed-forward
neural network with~$n$ layers
\begin{equation}
 \begin{aligned}
  \zv_1 & = \relu(\Dv_1^T\xv - b_1)       \\
  \zv_2 & = \relu(\Dv_2^T\zv_1 - b_2)     \\
        & \quad \vdots                    \\
  \zv_n & = \relu(\Dv_n^T\zv_{n-1} - b_n) \\
 \end{aligned}
\end{equation}
as executing a sequence of single-iteration ISTA, serving as an approximate
solution to the multi-layer sparse coding problem: find~$\{\zv_i\}_{i=1}^n$,
such that
\begin{equation}
 \begin{aligned}
  \xv = \Dv_1\zv_1       & , \quad \Vert \zv_1 \Vert_0 < \lambda_1, \zv_1 \ge 0, \\
  \zv_1 = \Dv_2\zv_2     & , \quad \Vert \zv_2 \Vert_0 < \lambda_2, \zv_2 \ge 0, \\
  \vdots \quad \quad     & , \quad \quad \vdots                                  \\
  \zv_{n-1} = \Dv_n\zv_n & , \quad \Vert \zv_n \Vert_0 < \lambda_n, \zv_n \ge 0, \\
 \end{aligned}
\end{equation}
where the bias terms~$\{b_i\}_{i=1}^n$ (in a similar manner to $\tau$) are
related to~$\{\lambda_i\}_{i=1}^n$, adjusting the sparsity of recovered code.
Furthermore, they reinterpret back-propagating through the deep neural network
as learning the dictionaries~$\{\Dv_i\}_{i=1}^n$. This connection offers a novel
breakthrough for understanding DNNs. In this paper, we extend this to the block
sparse scenario and apply it to solving our \nrsfm problem.

\begin{figure*}[t]
 \centering
 \includegraphics[width=\linewidth]{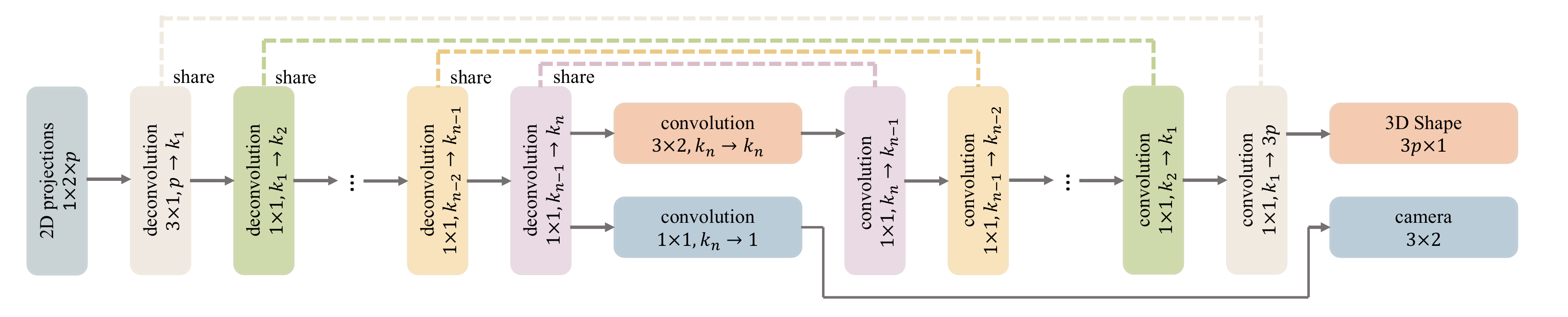}

 \caption{Deep \nrsfm architecture. The network can be divided into two parts:
  encoder and decoder that are symmetric and share convolution kernels (\ie
  dictionaries). The symbol~$a\times b, c \to d$ refers to the operator using
  kernel size~$a\times b$ with~$c$ input channels and $d$~output channels. }
 \label{fig:architecture}
\end{figure*}

\section{Deep Non-Rigid Structure from Motion}
\vspace{-0.25cm}
Under orthogonal projection, \nrsfm deals with the problem of factorizing
a 2D projection matrix $\Wv\in\RR^{p\times 2}$ as the product of a 3D shape
matrix $\Sv\in\RR^{p\times 3}$ and camera matrix $\Mv\in\RR^{3\times 2}$.
Formally,
\begin{equation}
 \Wv = \Sv\Mv,
 \label{eq:proj}
\end{equation}
\begin{equation}
 \Wv = \begin{bmatrix} u_1 & v_1 \\ u_2 & v_2 \\ \vdots & \vdots \\ u_p & v_p \end{bmatrix},~
 \Sv = \begin{bmatrix} x_1 & y_1 & z_1 \\ x_2 & y_2 & z_2 \\ \vdots & \vdots & \vdots \\ x_p & y_p & z_p \end{bmatrix},~
 \Mv^T\Mv = \Iv_2,
\end{equation}
where $(u_i, v_i), (x_i, y_i, z_i)$ are the image and world coordinates of the
$i$-th point. The goal of \nrsfm is to recover simultaneously
the shape~$\Sv$ and the camera $\Mv$ for each projection $\Wv$ in a given set
$\mathbb{W}$ of 2D landmarks. In a general \nrsfm including \sfc, this
set~$\mathbb{W}$ could contain deformations of a non-rigid object or various
instances from an object category.

\subsection{Modeling via multi-layer sparse coding}
\label{sec:mlscm}
To alleviate the ill-posedness of \nrsfm and also guarantee sufficient freedom
on shape variation, we propose a novel prior assumption on 3D shapes via
multi-layer sparse coding: The vectorization of $\Sv$ satisfies
\begin{equation}
 \begin{aligned}
  \sv = \Dv_1\psiv_1         & , \quad \Vert \psiv_1 \Vert_0 < \lambda_1, \psiv_1 \ge 0, \\
  \psiv_1 = \Dv_2\psiv_2     & , \quad \Vert \psiv_2 \Vert_0 < \lambda_2, \psiv_2 \ge 0, \\
  \vdots \quad \quad         & , \quad \quad \vdots                                      \\
  \psiv_{n-1} = \Dv_n\psiv_n & , \quad \Vert \psiv_n \Vert_0 < \lambda_n, \psiv_n \ge 0, \\
 \end{aligned}
 \label{eq:mlsc}
\end{equation}
where $\Dv_1 \in \RR^{3p\times k_1}, \Dv_2 \in \RR^{k_1 \times k_2}, \dots,
 \Dv_n \in \RR^{k_{n-1}\times k_n}$ are hierarchical dictionaries.
In this prior, each non-rigid shape is represented by a sequence of hierarchical
dictionaries and corresponding sparse codes. Each sparse code is determined by
its lower-level neighbor and affects the next-level. Clearly this hierarchy
adds more parameters, and thus more freedom into the system. We now show that
it paradoxically results in a more constrained global dictionary and sparse
code recovery.

\subsubsection{More constrained code recovery} In a classical single dictionary
system, the constraint on the representation is element-wise sparsity. Further,
the quality of its recovery entirely depends on the quality of the dictionary.
In our multi-layer sparse coding model, the optimal code not only
minimizes the difference between measurements~$\sv$ and~$\Dv_1\psiv_1$ along
with sparsity regularization~$\Vert \psiv_1 \Vert_0$, but also satisfies
constraints from its subsequent representations. This additional joint
inference imposes more constraints on code recovery, helps to control the
uniqueness and therefore alleviates its heavy dependency on the dictionary
quality.

\subsubsection{More constrained dictionary}
When all equality constraints are satisfied, the multi-layer sparse coding
model degenerates to a single dictionary system. From Equation~\ref{eq:mlsc},
by denoting~$\Dv^{(l)} = \prod_{i=1}^l \Dv_i$, it is implied that
$\sv = \Dv_1\Dv_2\dots\Dv_n\psiv_n = \Dv^{(n)}\psiv_n$.
However, this differs from other single dictionary models~\cite{zhu2014complex,
 zhu2013convolutional, kong2016prior, kong2016sfc, zhou20153d} in terms that a
unique structure is imposed on~$\Dv^{(n)}$~\cite{sulam2017multi}. The
dictionary~$\Dv^{(n)}$ is composed by simpler atoms hierarchically. For
example, each column of $\Dv^{(2)} = \Dv_1\Dv_2$ is a linear combination of
atoms in $\Dv_1$, each column of $\Dv^{(3)} = \Dv^{(2)}\Dv_3$ is a linear
combination of atoms in $\Dv^{(2)}$ and so on. Such a structure results in a
more constrained global dictionary and potentially leads to higher quality with
lower mutual coherence~\cite{donoho2006stable}.

\subsection{Multi-layer block sparse coding}
Given the proposed multi-layer sparse coding model, we now build a conduit
from the proposed shape code~$\{\psiv_i\}_{i=1}^k$ to the 2D projected points.
From Equation~\ref{eq:mlsc}, we reshape vector~$\sv$ to a matrix~$\Sv\in\RR^{p\times3}$ such that $\Sv = \Dv^\sharp_1(\psiv_1 \otimes \Iv_3)$,
where $\otimes$ is Kronecker product and $\Dv_1^\sharp\in\RR^{p\times 3k_1}$ is
a reshape of $\Dv_1$~\cite{dai2014simple}.
From linear algebra, it is well known that~$\Av\Bv \otimes \Iv = (\Av\otimes \Iv)(\Bv\otimes\Iv)$
given three matrices~$\Av, \Bv$, and identity matrix $\Iv$. Based on this lemma, we can derive
that
\begin{equation}
 \small
 \begin{aligned}
  \Sv = \Dv^\sharp_1(\psiv_1 \otimes \Iv_3)                               & , \quad \Vert \psiv_1 \Vert_0 < \lambda_1, \psiv_1 \ge 0, \\
  \psiv_1\otimes \Iv_3 = (\Dv_2 \otimes \Iv_3)(\psiv_2\otimes \Iv_3)      & , \quad \Vert \psiv_2 \Vert_0 < \lambda_2, \psiv_2 \ge 0, \\
  \vdots \quad \quad                                                      & , \quad \quad \vdots                                      \\
  \psiv_{n-1} \otimes \Iv_3 = (\Dv_n \otimes \Iv_3)(\psiv_n\otimes \Iv_3) & , \quad \Vert \psiv_n \Vert_0 < \lambda_n, \psiv_n \ge 0. \\
 \end{aligned}
 \label{eq:mlbsc_s}
\end{equation}

Further, from Equation~\ref{eq:proj}, by right multiplying the camera
matrix~$\Mv\in\RR^{3\times2}$ to the both sides of Equation~\ref{eq:mlbsc_s}
and denote $\Psiv_i = \psiv_i \otimes \Mv$, we obtain that
\begin{equation}
 \small
 \begin{aligned}
  \Wv = \Dv^\sharp_1 \Psiv_1                 & , \quad \Vert \Psiv_1 \Vert_0^{(3\times 2)} < \lambda_1, \\
  \Psiv_1 = (\Dv_2 \otimes \Iv_3)\Psiv_2     & , \quad \Vert \Psiv_2 \Vert_0^{(3\times 2)} < \lambda_2, \\
  \vdots \quad \quad                         & , \quad \quad \vdots                                     \\
  \Psiv_{n-1} = (\Dv_n \otimes \Iv_3)\Psiv_n & , \quad \Vert \Psiv_n \Vert_0^{(3\times 2)} < \lambda_n, \\
 \end{aligned}
 \label{eq:mlbsc_w}
\end{equation}
where $\Vert \cdot \Vert_0^{(3\times 2)}$ divides the argument matrix into
blocks with size $3\times 2$ and counts the number of active blocks.
Since~$\psiv_i$ has active elements less than $\lambda_i$, $\Psiv_i$ has active
blocks less than $\lambda_i$, that is $\Psiv_i$ is block sparse.
This derivation demonstrates that if the shape vector $\sv$ satisfies the
multi-layer sparse coding prior described by Equation~\ref{eq:mlsc}, then its
2D projection~$\Wv$ must be in the format of multi-layer \emph{block} sparse
coding described by Equation~\ref{eq:mlbsc_w}. We hereby interpret \nrsfm
as a hierarchical \emph{block} sparse dictionary learning problem \ie
factorizing $\Wv$ as products of hierarchical dictionaries~$\{\Dv_i\}_{i=1}^n$
and block sparse coefficients~$\{\Psiv_i\}_{i=1}^n$.

\subsection{Block ISTA and DNNs solution}
\label{sec:architecture}

Before solving the multi-layer block sparse coding problem in Equation~\ref{eq:mlbsc_w},
we first consider the single-layer problem:
\begin{equation}
 \min_{\Zv} \Vert \Xv - \Dv\Zv\Vert_F^2 \quad \st~\Vert \Zv \Vert_{0}^{(3\times2)} < \lambda.
\end{equation}
Inspired by ISTA, we propose to solve this problem by iteratively executing
the following two steps:
\begin{gather}
 \Vv = \Zv^{[i]} - \alpha\Dv^T(\Dv\Zv^{[i]} - \Xv), \\
 \Zv^{[i+1]} = \argmin_{\Uv} \frac{1}{2}\Vert \Uv - \Vv \Vert^2_F + \tau\Vert\Uv\Vert_{F1}^{(3\times2)},
\end{gather}
where $\Vert \cdot \Vert_{F1}^{(3\times2)}$ is defined as the summation of
Frobenius norm of each $3\times2$ block, serving as a convex relaxation of
block sparsity constraint. It is derived in~\cite{deng2013group} that the
second step has a closed-form solution computing each block separately by
$\small\Zv^{[i+1]}_j = (h_\tau(\Vert\Vv_j\Vert_F)/\Vert \Vv_j \Vert_F)\Vv_j$,
where the subscript $j$ represents the $j$-th block and $h_\tau$ is a soft
thresholding operator. However, soft thresholding the Frobenius norms for every
block brings unnecessary computational complexity. We show in
the supplementary material that an efficient approximation is
$\Zv^{[i+1]}_j = h_{b_j}(\Vv_j)$, where $b_j$ is the threshold for the $j$-th
block, controlling its sparsity.
Based on this approximation, a single-iteration block ISTA with step size
$\alpha=1$ can be represented by :
\begin{equation}
 \Zv = h_{\bv} \big(\Dv^T\Xv\big) = \relu(\Dv^T\Xv - \bv\otimes\one_{3\times 2}),
 \label{eq:singleBISTA}
\end{equation}
where $h_{\bv}$ is a soft thresholding operator using the $j$-th element $b_j$
as threshold of the $j$-th block and the second equality holds if $\Zv$ is
non-negative.

\subsubsection{Encoder}
Recall from Section~\ref{sec: scdnn} that the feed-forward pass through a deep
neural network can be considered as a sequence of single ISTA iterations and
thus provides an approximate recovery of multi-layer sparse codes. We follow
the same scheme: we first assume the multi-layer block sparse coding to be
non-negative and then sequentially use single-iteration block ISTA to solve it \ie
\begin{equation}
 \begin{aligned}
  \Psiv_1 & = \relu((\Dv^\sharp_1)^T\Wv - \bv_1\otimes\one_{3\times 2}),                \\
  \Psiv_2 & = \relu((\Dv_2 \otimes \Iv_3)^T\Psiv_1 - \bv_2\otimes\one_{3\times 2}),     \\
          & \quad \vdots                                                                \\
  \Psiv_n & = \relu((\Dv_n \otimes \Iv_3)^T\Psiv_{n-1} - \bv_n\otimes\one_{3\times 2}), \\
 \end{aligned}
\end{equation}
where thresholds $\bv_1, ..., \bv_n$ are learned, controlling the block sparsity.
This learning is crucial because in previous \nrsfm algorithms utilizing
low-rank~\cite{dai2014simple}, subspaces~\cite{zhu2014complex}
or compressible~\cite{kong2016prior} priors, the weight given to this prior
(\eg rank or sparsity) is hand-selected through a cumbersome cross validation
process. In our approach, this weighting is learned simultaneously with all
other parameters removing the need for any irksome cross validation process.
This formula composes the encoder of our proposed DNN.

\subsubsection{Decoder}
Let us for now assume that we can extract camera $\Mv$ and regular sparse
hidden code $\psiv_n$ from $\Psiv_n$ by some functions \ie $\Mv = \Fc(\Psiv_n)$
and $\psiv_n = \Gc(\Psiv_n)$, which will be discussed in the next section. Then
we can compute the 3D shape vector $\sv$ by:
\begin{equation}
 \begin{aligned}
  \psiv_{n-1} & = \relu(\Dv_n \psiv_n - \bv_n'), \\
              & \quad \vdots                     \\
  \psiv_1     & = \relu(\Dv_2 \psiv_2 - \bv_2'), \\
  \sv         & = \Dv^\sharp_1\psiv_1,
 \end{aligned}
\end{equation}
Note we preserve the ReLU and bias term during decoding to further enforce
sparsity and improve robustness. These portion forms the decoder of our DNN.


\subsubsection{Variation of implementation}
The Kronecker product of identity matrix $\Iv_3$ dramatically increases the
time and space complexity of our approach. To eliminate it and make parameter
sharing easier in modern deep learning environments~(\eg TensorFlow, PyTorch),
we reshape the filters and
features and show that the matrix multiplication in each step of the encoder
and decoder can be equivalently computed via multi-channel $1\times1$
convolution~($*$) and transposed convolution~($*^T$) \ie
\begin{equation}
 (\Dv_1^\sharp)^T\Wv = \dsf_1^\sharp *^T \wsf,
\end{equation}
where {\small $\dsf_1^\sharp \in \RR^{3\times1\times k_1 \times p},
   \wsf\in\RR^{1\times2\times p}$}\footnote{The filter dimension is
 height$\times$width$\times$\# of input channel$\times$\# of output channel.
 The feature dimension is height$\times$width$\times$\# of channel.}.
\begin{equation}
 (\Dv_{i+1}\otimes\Iv_3)^T\Psiv_i = \dsf_{i+1} *^T \Psi_{i},
\end{equation}
where {\small$\dsf_{i+1} \in \RR^{1\times1\times k_{i+1} \times k_i},
   \Psi_i\in\RR^{3\times2\times k_i}.$ }
\begin{equation}
 \Dv_i\psiv_i = \dsf_{i} * \psi_{i},
\end{equation}
where {\small$\dsf_{i} \in \RR^{1\times1\times k_{i} \times k_{i-1}},
   \psi_i\in\RR^{1\times1\times k_i}.$}

\subsubsection{Code and camera recovery}
Estimating~$\psiv_n$ and~$\Mv$ from~$\Psiv_n$ is discussed
in~\cite{kong2016prior} and solved by a closed-form formula. Due to its
differentiability, we could insert the solution directly within our pipeline.
An alternative solution is using an approximation \ie a fully connected layer
connecting $\Psiv_n$ and $\psiv_n$ and a linear combination among each blocks
of $\Psiv_n$ to estimate~$\Mv$, where the fully connected layer parameters and
combination coefficients are learned from data. In our experiments, we use the
approximate solution and represent them via convolutions, as shown in
Figure~\ref{fig:architecture}, for conciseness and maintaining proper dimensions.
Since the approximation has no way to force the orthonormal constraint on the
camera, we seek help from the loss function.

\subsubsection{Loss function} The loss function must measure the
reprojection error between input 2D points $\Wv$ and reprojected 2D points
$\Sv\Mv$ while simultaneously encouraging orthonormality of the estimated
camera~$\Mv$. One solution is to use spectral norm regularization of~$\Mv$
because spectral norm minimization is the tightest convex relaxation of the
orthonormal constraint~\cite{zhou20153d}. An alternative solution is to hard
code the singular values of~$\Mv$ to be exact ones with the help of Singular
Value Decomposition~(SVD). Even though SVD is generally non-differentiable, the
numeric computation of SVD is differentiable and most deep learning packages
implement its gradients~(\eg PyTorch, TensorFlow). In our implementation and
experiments, we use SVD to ensure the success of the orthonormal constraint and
a simple Frobenius norm to measure reprojection error,
\begin{equation}
 Loss = \Vert \Wv - \Sv\tilde{\Mv} \Vert_F, \quad \tilde{\Mv} = \Uv\Vv^T,
\end{equation}
where $\Uv\Sigmav\Vv^T = \Mv$ is the SVD of the camera matrix.

\begin{table}[]
\footnotesize
\begin{tabular}{
>{\columncolor[HTML]{EFEFEF}}r cccc
>{\columncolor[HTML]{EFEFEF}}c 
>{\columncolor[HTML]{EFEFEF}}c }
\hline\hline
Furnitures & \cellcolor[HTML]{EFEFEF}Bed & \cellcolor[HTML]{EFEFEF}Chair & \cellcolor[HTML]{EFEFEF}Sofa & \cellcolor[HTML]{EFEFEF}Table & Mean        & Relative      \\\hline
KSTA~\cite{gotardo2011kernel}       & 0.069                       & 0.158                         & 0.066                        & 0.217                         & 0.128          & 12.19         \\
BMM~\cite{dai2014simple}        & 0.059                       & 0.330                         & 0.245                        & 0.211                         & 0.211          & 20.12         \\
CNR~\cite{lee2016consensus}        & 0.227                       & 0.163                         & 0.835                        & 0.186                         & 0.352          & 33.55         \\
NLO~\cite{del2007non}        & 0.245                       & 0.339                         & 0.158                        & 0.275                         & 0.243          & 23.18         \\
RIKS~\cite{hamsici2012learning}       & 0.202                       & 0.135                         & 0.048                        & 0.218                         & 0.117          & 11.13         \\
SPS~\cite{kong2016prior}        & 0.971                       & 0.946                         & 0.955                        & 0.280                         & 0.788          & 74.96         \\
SFC~\cite{kong2016sfc}        & 0.247                       & 0.195                         & 0.233                        & 0.193                         & 0.217          & 20.67         \\\hline
OURS       & \textbf{0.004}              & \textbf{0.019}                & \textbf{0.005}               & \textbf{0.012}                & \textbf{0.010} & \textbf{1.00} \\\hline\hline
\end{tabular}
\caption{Quantitative comparison against state-of-the-art algorithms using IKEA dataset in normalized 3D error.}
\label{tab:sfc}
\end{table}

\begin{table*}[]
\footnotesize
\begin{tabular}{
>{\columncolor[HTML]{EFEFEF}}c c
>{\columncolor[HTML]{EFEFEF}}c c
>{\columncolor[HTML]{EFEFEF}}c c
>{\columncolor[HTML]{EFEFEF}}c c
>{\columncolor[HTML]{EFEFEF}}c c
>{\columncolor[HTML]{EFEFEF}}c c
>{\columncolor[HTML]{EFEFEF}}c c
>{\columncolor[HTML]{EFEFEF}}c c
>{\columncolor[HTML]{EFEFEF}}c c
>{\columncolor[HTML]{EFEFEF}}c }
\hline
\hline
Methods     & \multicolumn{2}{c}{\cellcolor[HTML]{EFEFEF}KSTA~\cite{gotardo2011kernel}} & \multicolumn{2}{c}{\cellcolor[HTML]{EFEFEF}BMM~\cite{dai2014simple}} & \multicolumn{2}{c}{\cellcolor[HTML]{EFEFEF}CNS~\cite{lee2016consensus}} & \multicolumn{2}{c}{\cellcolor[HTML]{EFEFEF}MUS~\cite{agudo2018image}} & \multicolumn{2}{c}{\cellcolor[HTML]{EFEFEF}NLO~\cite{del2007non}} & \multicolumn{2}{l}{\cellcolor[HTML]{EFEFEF}RIKS~\cite{hamsici2012learning}} & \multicolumn{2}{c}{\cellcolor[HTML]{EFEFEF}SPS~\cite{kong2016prior}} & \multicolumn{2}{c}{\cellcolor[HTML]{EFEFEF}SFC~\cite{kong2016sfc}} & \multicolumn{2}{c}{\cellcolor[HTML]{EFEFEF}OURS} \\ \hline
Aeroplane   & 0.145                   & 0.175                  & 0.843                  & 1.459                  & 0.263                  & 0.416                  & 0.261                    & -                    & -                    & 0.876                    & -                     & 0.132                    & -                    & 0.930                    & -                    & 0.504                    & -                & \textbf{0.024}                \\
Bicycle     & 0.442                   & 0.245                  & 0.308                  & 1.376                  & -                      & 0.356                  & 0.178                    & -                    & -                    & 0.269                    & -                     & 0.136                    & -                    & 1.322                    & -                    & 0.372                    & -                & \textbf{0.003}                \\
Bus         & 0.214                   & 0.199                  & 0.300                  & 1.023                  & -                      & 0.250                  & 0.113                    & -                    & -                    & 0.140                    & -                     & 0.160                    & -                    & 0.604                    & -                    & 0.251                    & -                & \textbf{0.004}                \\
Car         & 0.159                   & 0.152                  & 0.266                  & 1.278                  & 0.099                  & 0.258                  & 0.078                    & -                    & -                    & 0.104                    & -                     & 0.097                    & -                    & 0.872                    & -                    & 0.282                    & -                & \textbf{0.009}                \\
Chair       & 0.399                   & 0.186                  & 0.357                  & 1.297                  & -                      & 0.170                  & 0.210                    & -                    & -                    & 0.146                    & -                     & 0.192                    & -                    & 1.046                    & -                    & 0.226                    & -                & \textbf{0.007}                \\
Diningtable & 0.372                   & 0.267                  & 0.422                  & 1.00                   & -                      & 0.170                  & 0.264                    & -                    & -                    & 0.109                    & -                     & 0.207                    & -                    & 1.050                    & -                    & 0.221                    & -                & \textbf{0.060}                \\
Motorbike   & 0.270                   & 0.255                  & 0.336                  & 0.857                  & -                      & 0.457                  & 0.222                    & -                    & -                    & 0.432                    & -                     & 0.118                    & -                    & 0.986                    & -                    & 0.361                    & -                & \textbf{0.002}                \\
Sofa        & 0.298                   & 0.307                  & 0.279                  & 1.126                  & 0.214                  & 0.250                  & 0.167                    & -                    & -                    & 0.149                    & -                     & 0.228                    & -                    & 1.328                    & -                    & 0.302                    & -                & \textbf{0.004}                \\\hline
Average     & 0.287                   & 0.223                  & 0.388                  & 1.178                  & 0.192                  & 0.291                  & 0.186                    & -                    & -                    & 0.278                    & -                     & 0.159                    & -                    & 1.017                    & -                    & 0.315                    & -                & \textbf{0.014}                \\
Relative    & -                       & 15.33                  & -                      & 80.76                  & -                      & 19.95                  & -                        & -                    & -                    & 19.09                    & -                     & 10.92                    & -                    & 69.74                    & -                    & 21.61                    & -                & \textbf{1.00}                \\\hline \hline
Aeroplane   & 0.183                   & 0.207                   & 0.566                     & 1.465                    & 0.294                   & 0.460                  & 0.271                     & -                    & -                     & 0.758                    & -                     & 0.146                     & -                     & 0.888                    & -                     & 0.521                    & -                 & \textbf{0.032}                \\
Bicycle     & 0.457                   & 0.232                   & 0.307                     & 1.404                    & -                       & 0.359                  & 0.188                     & -                    & -                     & 0.275                    & -                     & 0.139                     & -                     & 0.851                    & -                     & 0.379                    & -                 & \textbf{0.007}                \\
Bus         & 0.218                   & 0.197                   & 0.255                     & 0.764                 & -                       & 0.264                  & 0.122                     & -                    & -                     & 0.141                    & -                     & 0.159                     & -                     & 1.110                    & -                     & 0.264                    & -                 & \textbf{0.021}                \\
Car         & 0.164                   & 0.139                   & 0.161                     & 1.744                 & 0.122                   & 0.265                  & 0.093                     & -                    & -                     & 0.105                    & -                     & 0.102                     & -                     & 0.804                    & -                     & 0.281                    & -                 & \textbf{0.010}                \\
Chair       & 0.396                   & 0.203                   & 0.258                     & 1.197                 & -                       & 0.171                  & 0.220                     & -                    & -                     & 0.145                    & -                     & 0.193                     & -                     & 1.016                    & -                     & 0.223                    & -                 & \textbf{0.017}                \\
Diningtable & 0.383                   & 0.249                   & 0.358                     & 1.105                 & -                       & 0.172                  & 0.267                     & -                    & -                     & 0.114                    & -                     & 0.227                     & -                     & 1.213                    & -                     & 0.222                    & -                 & \textbf{0.034}                \\
Motorbike   & 0.290                   & 0.227                   & 0.299                     & 1.117                 & -                       & 0.459                  & 0.233                     & -                    & -                     & 0.254                    & -                     & 0.125                     & -                     & 0.915                    & -                     & 0.351                    & -                 & \textbf{0.011}                \\
Sofa        & 0.294                   & 0.436                   & 0.240                     & 1.143                 & 0.228                   & 0.255                  & 0.174                     & -                    & -                     & 0.152                    & -                     & 0.239                     & -                     & 1.164                    & -                     & 0.306                    & -                 & \textbf{0.008}                \\ \hline
Average     & 0.298                   & 0.236                   & 0.305                     & 1.232                 & 0.215                   & 0.300                  & 0.196                     & -                    & -                     & 0.243                    & -                     & 0.166                     & -                     & 0.995                    & -                     & 0.318                    & -                 & \textbf{0.017}                \\
Relative    & -                       & 16.90                   & -                         & 88.78                 & -                       & 21.49                  & -                         & -                    & -                     & 17.39                    & -                     & 11.90                     & -                     & 71.11                    & -                     & 22.78                    & -                 & \textbf{1.27}                 \\ \hline
\end{tabular}
\caption{Quantitative evaluation on PASCAL3D+ dataset. We conduct experiments on both original and 
noisy 2D annotations, listed at the upper and lower half of table respectively.
The symbol `-' indicates either algorithm implementation or data is missing. The shaded columns are erros using our processed data and others are copied from Table 2 in~\cite{agudo2018image}. 
Relative errors are computed with respect to our method, the most accurate solution, without 
noise perturbation.
Our data and implementation will be publicly accessible for future comparison. 
}
\label{tab:pascal3d}
\end{table*}

\section{Experiments}
\vspace{-0.25cm}
We conduct extensive experiments to evaluate the performance of our deep
solution for solving \nrsfm and \sfc problems. For quantitative evaluation,
we follow the metric \ie normalized mean 3D error, reported 
in~\cite{akhter2009nonrigid, dai2014simple, gotardo2011kernel, agudo2018image}.
A detailed description of our architectures is in the supplementary material.
Our implementation and processed data will be publicly accessible for future
comparison.

\subsection{S\textbf{\textit{f}}C on IKEA furniture}
We first apply our method to a furniture dataset, IKEA dataset~\cite{lpt2013ikea,
 wu2016single}. The IKEA dataset contains four object categories: bed, chair,
sofa, and table. For each object category, we employ all annotated 2D point
clouds and augment them with 2K ones projected from the 3D ground-truth
using randomly generated orthonormal cameras\footnote{Augmentation is utilized
 due to limited valid frames, because the ground-truth cameras are
 partially missing.}. The error evaluated on real images are reported and summarized into
Table~\ref{tab:sfc}. One can observe that our method outperforms baselines
in the order of magnitude, clearly showing the superiority of our model. 
For qualitative evaluation, we randomly select a frame from each object category
and show them in Figure~\ref{fig:sfc} against ground truth and baselines. It shows 
that our reconstructed landmarks effectively depict the 3D geometry of objects
and our method is able to cover subtle geometric details.

\subsection{\sfc on PASCAL3D+}
We then apply our method to PASCAL3D+ dataset~\cite{xiang2014beyond} which contains 
twelve object categories and each category is labeled by approximately eight 3D CADs. 
To compare against more baselines, we follow the experiment setting reported 
in~\cite{agudo2018image} and use the same normalized 3D error metric. We report our errors
in Table~\ref{tab:pascal3d} emphasized by shading and concatenate the numbers copied 
from the Table 2 in~\cite{agudo2018image} for comparison.  Note that the errors are not 
exactly reproduced even though using the same dataset and algorithm implementation, because 
the data preparation details are missing. However, one can clearly see that our 
proposed method achieves extremely accurate reconstructions with more than ten times of 
smaller 3D error. This large margin makes the slight difference caused by data preparation 
even less noticeable. It clearly demonstrates the high precision of our proposed
deep neural network and also the superior robustness in noisy situations.

\begin{figure}[b]
 \centering
 \includegraphics[width=\linewidth]{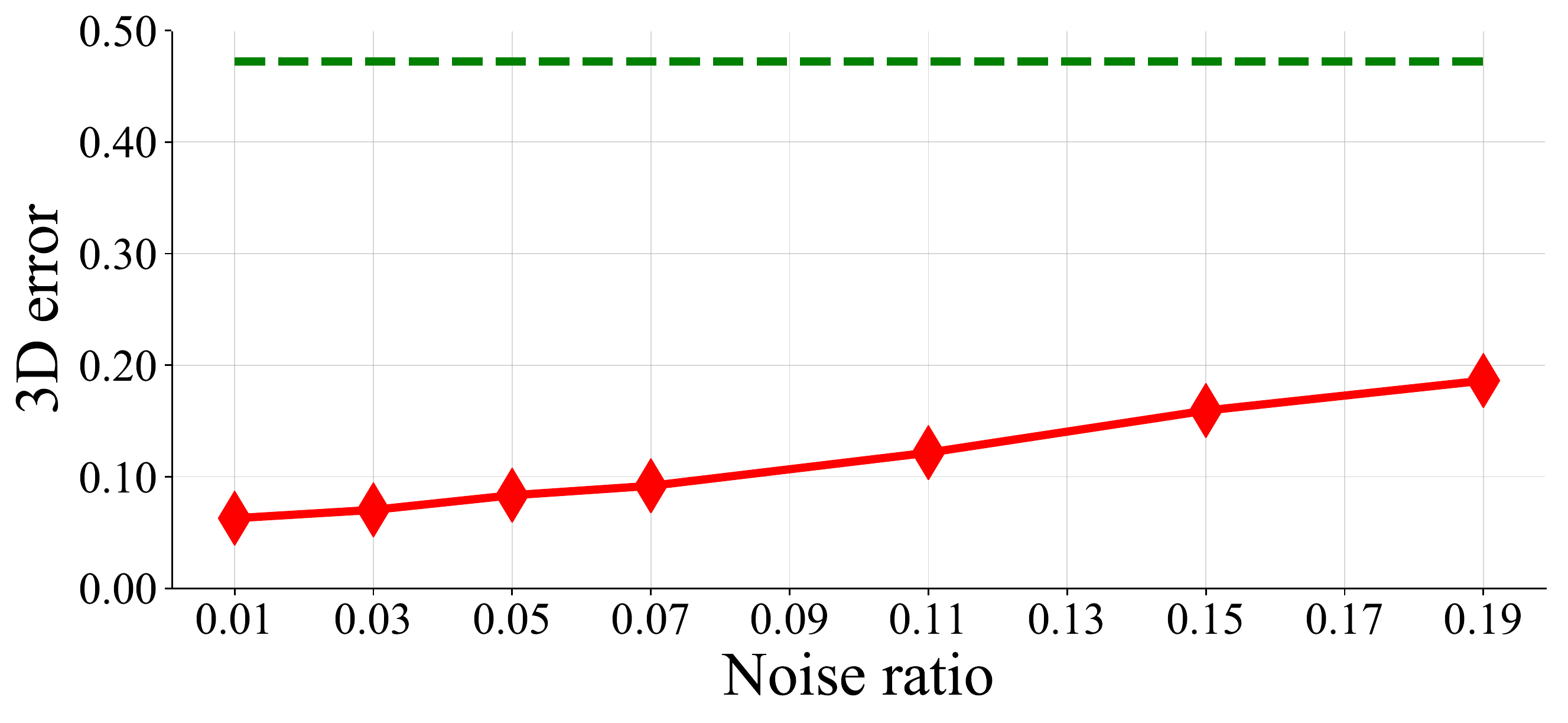}
 \caption{\nrsfm with noise perturbation. The red solid line is ours while the
  green dashed line is CNS~\cite{lee2016consensus}, the best performance of state-of-the-arts with \emph{no} noise perturbation.}
 \label{fig:noise}
\end{figure}

\begin{table*}[]
\begin{tabular}{
>{\columncolor[HTML]{EFEFEF}}r cccccccccc
>{\columncolor[HTML]{EFEFEF}}c 
>{\columncolor[HTML]{EFEFEF}}c }
\hline\hline
Subjects                                         & \cellcolor[HTML]{EFEFEF}01 & \cellcolor[HTML]{EFEFEF}05 & \cellcolor[HTML]{EFEFEF}18 & \cellcolor[HTML]{EFEFEF}23 & \cellcolor[HTML]{EFEFEF}64 & \cellcolor[HTML]{EFEFEF}70 & \cellcolor[HTML]{EFEFEF}102 & \cellcolor[HTML]{EFEFEF}106 & \cellcolor[HTML]{EFEFEF}123 & \cellcolor[HTML]{EFEFEF}127 & Average                                           & Relative                                         \\\hline
CNS~\cite{lee2016consensus}                                                & 0.613                      & 0.657                      & 0.541                      & 0.603                      & 0.543                      & 0.472                      & 0.581                       & 0.636                       & 0.479                       & 0.644                       & 0.577                                             & 5.66                                             \\
NLO~\cite{del2007non}                                                & 1.218                      & 1.160                      & 0.917                      & 0.998                      & 1.218                      & 0.836                      & 1.144                       & 1.016                       & 1.009                       & 1.050                       & 1.057                                             & 10.37                                            \\
SPS~\cite{kong2016prior}                                                & 1.282                      & 1.122                      & 0.953                      & 0.880                      & 1.119                      & 1.009                      & 1.078                       & 0.957                       & 0.828                       & 1.021                       & 1.025                                             & 10.06                                            \\\hline
OURS                                               & \textbf{0.175}             & \textbf{0.220}             & \textbf{0.081}             & \textbf{0.053}             & \textbf{0.082}             & \textbf{0.039}             & \textbf{0.115}              & \textbf{0.113}              & \textbf{0.040}              & \textbf{0.095}              & \textbf{0.101}                                    & \textbf{1.00}                                       \\\hline
UNSEEN     & 0.362                      & 0.331                      & 0.437                      & 0.387                      & 0.174                      & 0.090                      & 0.413                       & 0.194                       & 0.091                       & 0.388                       & 0.287          & 2.81      \\\hline\hline
\end{tabular}
 \caption{Quantitative comparison on solving large-scale \nrsfm problem using CMU MoCap dataset. Each subject contains more than ten thousand of frames. Due to huge volume of frames, KSTA~\cite{gotardo2011kernel}, BMM~\cite{dai2014simple}, 
MUS~\cite{agudo2018image}, RIKS~\cite{hamsici2012learning} all fail and thus are
omitted in the table. UNSEEN refers to the errors of the motions that
 are not accessible during training. This is used to demonstrate the well generalization of our proposed
 network, which is especially important in real world applications.}
 \label{tab:nrsfm}
\end{table*}

\subsection{Large-scale NRS\textbf{\textit{f}}M on CMU MoCap}
\begin{figure*}[t]
 \centering
 \includegraphics[width=\linewidth]{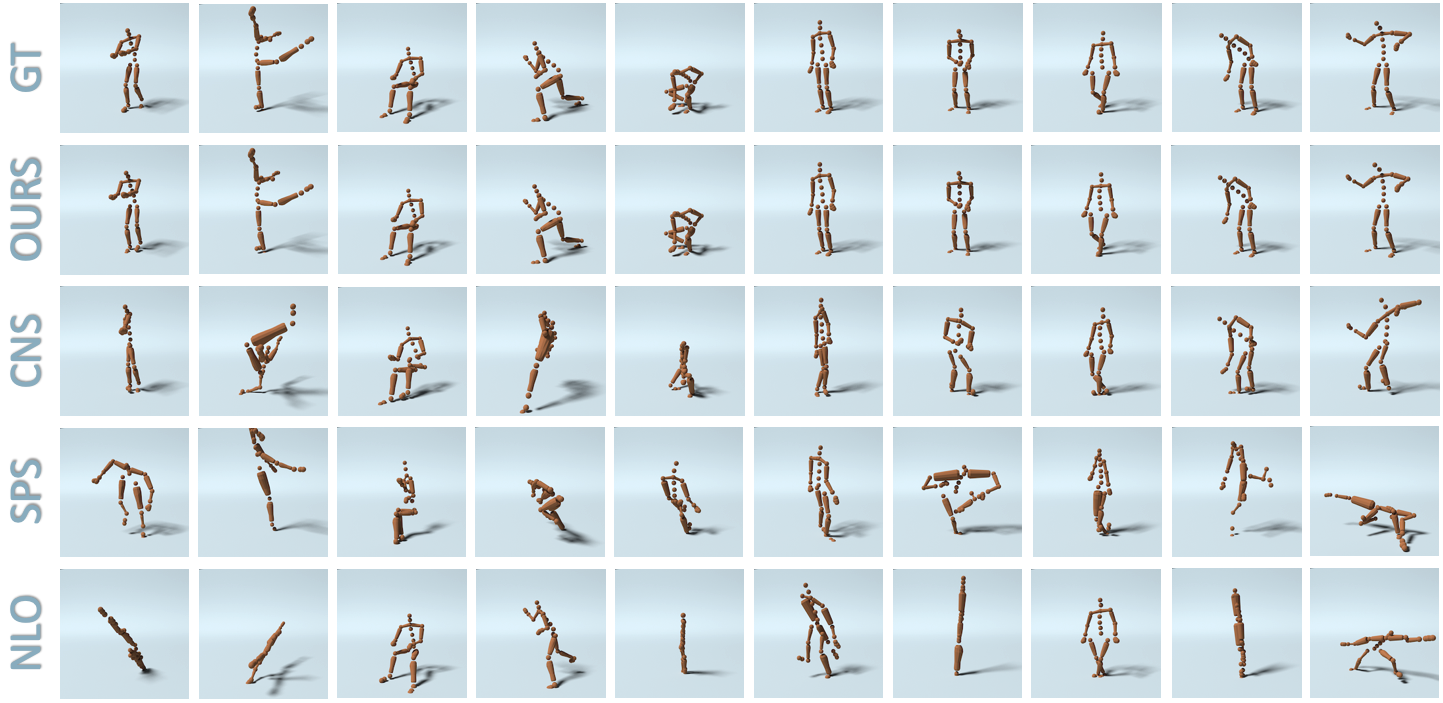}

 \caption{Qualitative evaluation on CMU Mocap dataset. Top to bottom: ground-truth, ours,
 CNS~\cite{lee2016consensus}, SPS~\cite{kong2016prior}, NLO~\cite{del2007non}. Each 
 column corresponds to reconstructions of a certain frame, randomly selected from
 each subject. Spheres are reconstructed landmarks while bars are for visualization. 3D shapes are 
 already aligned to the ground truth by orthonormal matrix.}
 \label{fig:nrsfm}

 \vspace{5mm}

  \includegraphics[width=\linewidth]{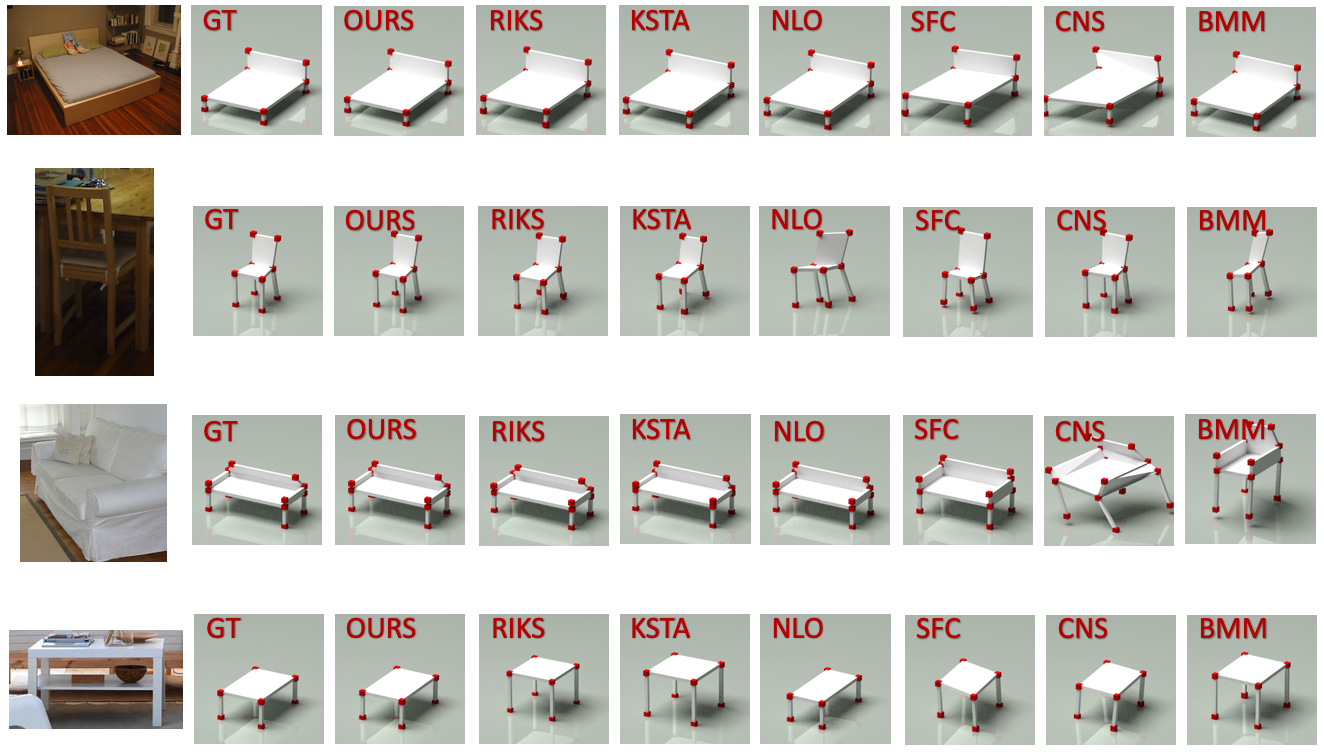}

 \caption{Qualitative evaluation on IKEA dataset. Landmarks projected by annotated cameras are omitted from images. In each
 rendering, red cubes are reconstructed points while the planes and bars are manually added for 
 descent visualization. Left to right: annotated image, ground-truth, ours,
 RIKS~\cite{hamsici2012learning},
 KSTA~\cite{gotardo2011kernel}, NLO~\cite{del2007non}, SFC~\cite{kong2016sfc},
 CNS~\cite{lee2016consensus}, BMM~\cite{dai2014simple}.} 
 \label{fig:sfc}
\end{figure*}

We finally apply our method to solving the problem of \nrsfm using the CMU motion
capture dataset\footnote{http://mocap.cs.cmu.edu/}. We randomly select 10 subjects
out of 144 and for each subject we concatenate $80\%$ of motions to form large image 
collections and remain the left $20\%$ as unseen motions for testing generalization. 
Note that in this experiment, each subject contains more than ten thousands of frames. 
We compare our method against state-of-the-art methods, summarized in Table~\ref{tab:nrsfm}.
Due to huge volume of frames, KSTA~\cite{gotardo2011kernel}, BMM~\cite{dai2014simple}, 
MUS~\cite{agudo2018image}, RIKS~\cite{hamsici2012learning} all fail and thus are
omitted in the table. We also report the normalized 3D error on unseen motions,
labeled as UNSEEN. One can see that our method obtains impressive reconstruction 
performance and outperforms others again in every sequences. Moreover, our network
also show a well generalization to unseen data which improve the effectiveness in
real world applications. For qualitative evaluation, we randomly select a frame 
from each subject and render the reconstructed human skeleton in Figure~\ref{fig:nrsfm}. 
This visually verifies the impressive performance of our deep solution.

\subsubsection{Robustness analysis} To analyze the robustness of our method, we
re-train the neural network for Subject 70 using projected points with
Gaussian noise perturbation. The results are summarized in Figure~\ref{fig:noise}.
The noise ratio is defined as $\Vert \text{noise} \Vert_F / \Vert \Wv \Vert_F$.
One can see that the error increases slowly with adding higher magnitude of noise and
when adding up to $20\%$ noise to image coordinates, our method in red still achieves 
better reconstruction compared to the best baseline with no noise perturbation (in green). 
This experiment clearly demonstrates the robustness of our model and its high 
accuracy against state-of-the-art works.

\begin{figure}[b]
 \centering
 \includegraphics[width=\linewidth]{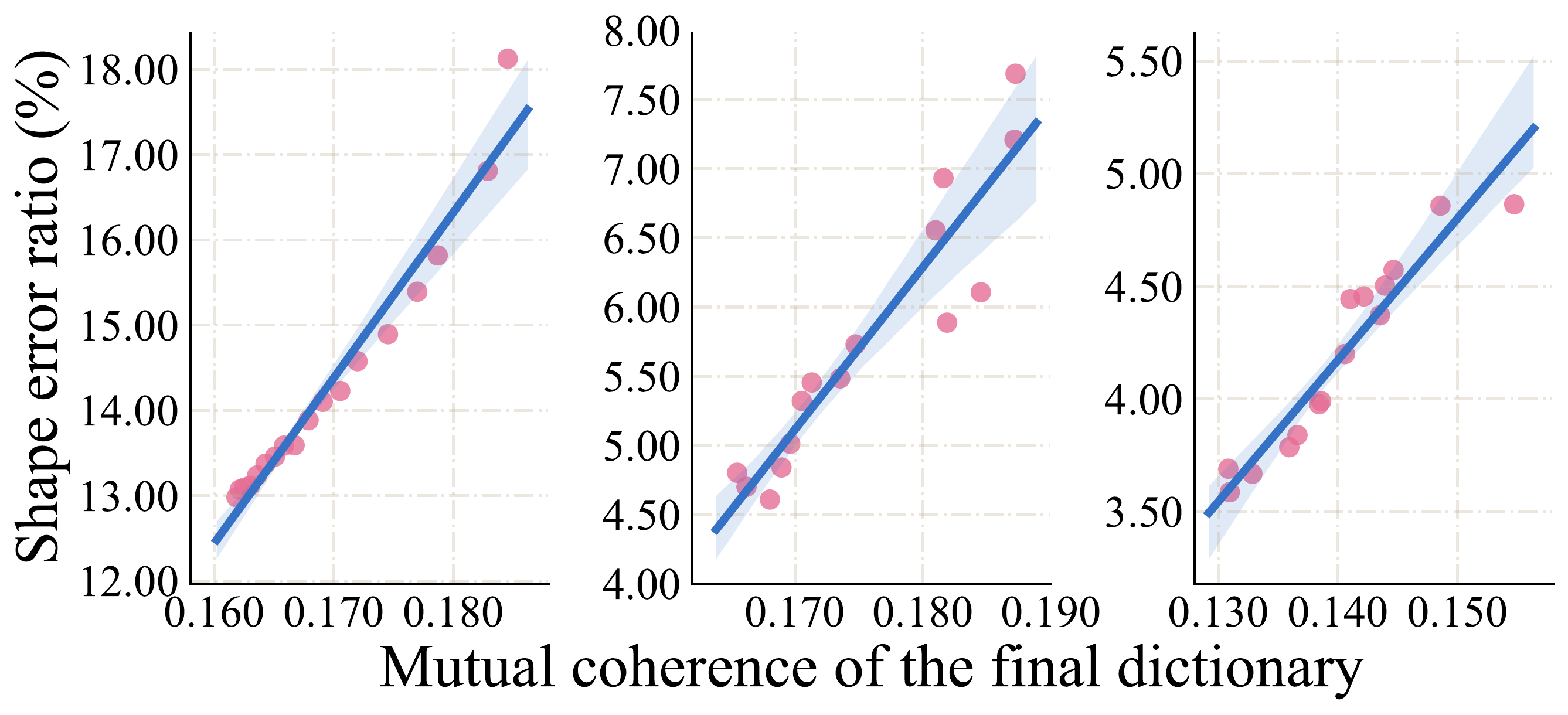}
 \caption{A scatter plot of the shape error ratio in percentage against the
  final dictionary coherence. A line is fitted based on the data. The left
  comes from subject 05, the middle from subject 18, the right from subject 64.}
 \label{fig:coherence}
\end{figure}

\subsubsection{Missing data}
Landmarks are not always visible from the camera owing to the occlusion by other
objects or itself. In the present paper, we focus on a complete measurement
situation not accounting for invisible landmarks. However, thanks to recent
progress in matrix completion, our method can be easily extended to missing data.
Moreover, in our experiments, we observe that deep neural network shows a well
tolerance of missing data. Simply setting missing 2D coordinates as zeros provides
satisfactory results. Such technique is widely used in deep-learning-based 
depth map reconstruction from sparse observations~\cite{chen2018estimating,
 mal2018sparse, li2018depth, liao2017parse, cadena2016multi}. These two solutions
make our central pipeline of DNN more easily to adapt to handling missing data.

\subsection{Coherence as guide}
As explained in Section~\ref{sec:mlscm}, every sparse code~$\psiv_i$ is
constrained by its subsequent representation and thus the quality of code
recovery depends less on the quality of the corresponding dictionary. However,
this is not applicable to the final code~$\psiv_n$, making it least constrained
with the most dependency on the final dictionary~$\Dv_n$. From this
perspective, the quality of the final dictionary measured by mutual
coherence~\cite{donoho2006stable} could serve as a lower bound of the entire
system. To verify this, we compute the error and coherence in a fixed interval
during training in \nrsfm experiments. We consistently observe strong
correlations between 3D reconstruction error and the mutual coherence of the
final dictionary. We plot this relationship in Figure~\ref{fig:coherence}. We
thus propose to use the coherence of the final dictionary as a measure of model
quality for guiding training to efficiently avoid over-fitting especially when
3D evaluation is not available. This improves the utility of our deep \nrsfm in
future applications without 3D ground-truth.

\section{Conclusion}
\vspace{-0.25cm}
In this paper, we proposed multi-layer sparse coding as a novel prior
assumption for representing 3D non-rigid shapes and designed an innovative
encoder-decoder neural network to solve the problem of \nrsfm using \emph{no} 3D
supervision. The proposed network was derived by generalizing the classical sparse
coding algorithm ISTA to a block sparse scenario. The proposed network architecture
is mathematically interpretable as solving a \nrsfm multi-layer sparse dictionary
learning problem. Extensive experiments demonstrated our superior performance
against the state-of-the-art methods and our generalization to
unseen data. Finally, we proposed to use the coherence of the final dictionary
as a model quality measure, offering a practical way to avoid over-fitting and
select the best checkpoint during training without relying on 3D ground-truth.

{\small
\bibliographystyle{ieee}
\bibliography{egbib}
}

\end{document}